\documentclass[a4paper,12pt]{article}

\usepackage{amsmath}
\usepackage{graphicx}

\usepackage{url}
\usepackage{epsfig}
\usepackage{mahnig}

\begin{document}
 \title{Computing factorized approximations of Pareto-fronts using mNM-landscapes and Boltzmann distributions}

\author{Roberto Santana, Alexander Mendiburu, Jose A. Lozano \\
Intelligent Systems Group. University of the Basque Country (UPV/EHU) \\
\{roberto.santana,alexander.mendiburu,ja.lozano\}@ehu.es}% <-this % stops a space

\date{}
\maketitle

\begin{abstract}
 NM-landscapes have been recently introduced as a class of tunable rugged models. They are a subset of the general interaction models where all the interactions are of order less or equal $M$. The Boltzmann distribution has been extensively applied in single-objective evolutionary algorithms to implement selection and study the theoretical properties of model-building algorithms. In this paper we propose the combination of the multi-objective NM-landscape model and the Boltzmann distribution to obtain Pareto-front approximations. We investigate the joint effect of the parameters of the NM-landscapes and the probabilistic factorizations in the shape of the Pareto front approximations. \\
{\bf{keywords}}:  multi-objective optimization, NM-landscape, factorizations, Boltzmann distribution
\end{abstract}

\section{Introduction}

 One important question in multi-objective evolutionary algorithms (MOEAs) is how the structure of the interactions between the variables of the problem influences the different objectives and impacts in the characteristics of the Pareto front (e.g. discontinuities, clustered structure, etc.). The analysis of interactions is also important because there is a class of MOEAs that explicitly capture and represent these interactions to make a more efficient search \cite{Bosman_and_Thierens:2002,Marti_et_al:2010,Pelikan_et_al:2006a}. In this paper, we approach this important question by combining the use of a multi-objective fitness landscape model with the definition of probability distributions on the search space and different factorized approximations to these joint distributions. Our work follows a similar methodology to the one used in  \cite{Muehlenbein_and_Mahnig:2002a,Muhlenbein_et_al:1999,Okabe_et_al:2004a,Santana_et_al:2005b} to investigate the relationship between additively decomposable single-objective functions and the performance of estimation of distribution algorithms (EDAs)  \cite{Larranaga_et_al:2012,Lozano_et_al:2005}.  

 Landscapes models are very useful to understand the behavior of optimizers under different hypothesis about the complexity of the fitness function. Perhaps the best known example of such models is the NK fitness landscape \cite{Kauffman:1993}, a parametrized model of a fitness landscape that allows to explore the way in which the neighborhood structure and the strength of interactions between neighboring variables determine the ruggedness of the landscape. One relevant aspect of the NK-fitness landscape is its simplicity and wide usability across disciplines from diverse domains. 

 Another recently introduced landscape model is the NM-landscape  \cite{Manukyan_et_al:2014}. It can be seen as a generalization of the NK-landscape. This model has a number of attributes that makes it particularly suitable to control the strength of the interactions between subsets of variables of different size.  In addition, it is not restricted to binary variables and allows the definition of functions on any arity. 

 In \cite{Santana_et_al:2015b}, the NM-landscape was extended to  multi-objective problems and used to study the influence of the parameters in the characteristics of the MOP. We build on the work presented in \cite{Santana_et_al:2015b} to propose the use of the multi-objective NM-landscape (mNM-landscape) for investigating how the patterns of interactions in the landscape model influence the shape of the Pareto front. We go one step further and propose the use of factorized approximations computed from the landscapes to approximate the Pareto fronts. We identify the conditions in which these approximations can be accurate.

% \cite{Buzas_and_Dinitz:2014}

%The paper is organized as follows: In the next section, the NM landscape is introduced and its properties are discussed.  Our approach to evolve MNK-landscape instances of high complexity is explained in Section~\ref{sec:MULTI-MN}. Section~\ref{sec:RELWORK} discusses research related to our proposal. Section~\ref{sec:EXPE} describes the experimental framework to evaluate our proposal and  presents the numerical results. The main contributions of the paper are summarized in Section~\ref{sec:CONCLU}, where some lines for future research are also discussed. 

\section{NM-landscape} \label{sec:NMLAND}

\subsection{Definition}

Let $ {\bf{X}}=(X_1,\ldots ,X_N)$ denote a vector of discrete variables. We will  use ${\bf{x}}=(x_1,\ldots ,x_N)$ to denote an assignment to the variables. $S$ will denote a set of indices in $\{1, \ldots, N\}$, and $X_S$ (respectively $x_S$)   a subset of the variables of ${\bf{X}}$ (respectively ${\bf{x}}$) determined by the indices in $S$.

A fitness landscape $F$ can be defined for $N$ features using a general parametric interaction model of the form:

\begin{equation}
 F({\bf{x}}) = \sum_{k=1}^{l} \beta_{U_k} \prod_{i \in U_k} x_i \label{eq:INTMODEL}
\end{equation}
where $l$ is the number of terms, and each of the $l$ coefficients $\beta_{U_k} \in \mathcal{R}$. For $k=1, \dots, l$, $U_k \subseteq \{1,2,\dots,N\}$, where $U_k$ is a set of indices of the features in the $k$th term, and the length $|U_k|$ is the order of the interaction. By convention \cite{Manukyan_et_al:2014}, it is assumed that when $U_k = \emptyset$, $\prod_{j \in U_k} x_j \equiv 1$. Also by convention, we assume that the model is  defined for binary variables represented as $x_i \in \{-1,1\}$. 

 The NM models \cite{Manukyan_et_al:2014} comprise the set of all general interactions models specified by Equation~\ref{eq:INTMODEL}, with the following constraints:

 \begin{itemize}
  \item All coefficients  $\beta_{U_k}$ are non-negative.
  \item Each feature value $x_i$ ranges from negative to positive values.
  \item The absolute value of the lower bound of the range is lower or equal than the upper bound of the range of $x_i$.
 \end{itemize}
    
One key element of the model is how the parameters of the interactions are generated. In \cite{Manukyan_et_al:2014}, each $\beta_{U_k}$ is generated from  $e^{-abs(\mathcal{N}(0,\sigma))}$, where $\mathcal{N}(0,\sigma)$ is a random number drawn from a Gaussian distribution with mean $0$ and standard deviation $\sigma$. Increasing $\sigma$ determines smaller range and increasing clumping of fitness values. In this paper, we use the same procedure to generate the $\beta_{U_k}$ parameters.

We will  focus on NM-models defined on the binary alphabet. In this case, the NM-landscape has a global maximum that is reached at ${\bf{x}} = (1,\dots, 1)$  \cite{Manukyan_et_al:2014}.

\section{Multi-objective NM-landscapes} \label{sec:MULTI-MN}

 The multi-objective NM-landscape model (mNM-landscape) is defined  \cite{Santana_et_al:2015b} as a vector function mapping binary vectors of solutions  into $m$ real numbers ${\bf{f(.)}}= (f_1(.),f_2(.),\dots,f_m(.)): \mathcal{B}^N \rightarrow \mathcal{R}^m$, where $N$ is the number of variables, $m$ is the number of objectives, $f_i(.)$ is the $i$-th objective function, and $\mathcal{B}=\{-1,1\}$. ${\bf{M}}=\{M_1,\dots,M_m\}$ is a set of integers where $M_i$ is the maximum order of the interaction in the $i$-th landscape. Each $f_i({\bf{x}})$ is defined similarly to Equation~\eqref{eq:INTMODEL} as:

\begin{equation}
  f_i({\bf{x}}) =  \sum_{k=1}^{l_i} \beta_{U_{k_i}} \prod_{j \in U_{k_i}} x_j,  
 \label{eq:MINTMODEL}
\end{equation}
where $l_i$ is the number of terms in objective $i$, and each of the $l_i$ coefficients $\beta_{U_{k_i}} \in \mathcal{R}$. For $k=1, \dots, l_i$, $U_{k_i} \subseteq \{1,2,\dots,N\}$, where $U_{k_i}$ is a set of indices of the features in the $k$th term, and the length $|U_{k_i}|$ is the order of the interaction. 

Notice that the mNM fitness landscape model allows that each objective may have a different maximum order of interactions. The mNM-landscape is inspired by previous extensions of the NK fitness landscape model to multi-objective functions \cite{Aguirre_and_Tanaka:2004,Aguirre_and_Tanaka:2007,Lopez_et_al:2014,Verel_et_al:2013}. 

%\subsection{mNM-landscapes with constraints} \label{sec:CONST}

  One of our goals is to use the mNM-landscape to investigate the effect that the combination of objectives with different structures of interactions has in the characteristics of the MOP. Without lack of generality, we will focus on bi-objective mNM-landscapes (i.e., $m=2$) and will establish some connections between the objectives. In this section we explain how the constrained mNM-landscapes are designed. 

% \subsubsection{Objectives with global optima reached at different points}

As previously explained, the NM-model is defined for $(x_1,\dots,x_N)\in \{-1,1\}$. However, we will use a representation in which  $(x_1,\dots,x_N)\in \{0,1\}$. The following transformation  \cite{Santana_et_al:2015b} maps the desired representation to the one used by the mNM-landscape. Given the analysis presented in  \cite{Manukyan_et_al:2014}, it also guarantees that the Pareto set will comprise at least two points, respectively reached at $(0,\dots,0)$ and $(1,\dots,1)$ for objectives $f_1$ and $f_2$.

\begin{align}
   f_1({\bf{y}}):& \;   y_i     = -2x_i+1&     \label{eq:TRANSF1}   \\              
   f_2({\bf{z}}):& \;   z_i     =  2x_i-1&     \label{eq:TRANSF2}
\end{align}
where ${\bf{y}} = (y_1,\dots,y_N) \in \{-1,1\}$  and ${\bf{z}} = (z_1,\dots,z_N)\in \{-1,1\}$ are the new variables obtained after the corresponding transformation have been applied to ${\bf{x}} = (x_1,\dots,x_N) \in \{0,1\}$.

 When the complete space of solutions is evaluated, we add two normalization steps to be able to compare landscapes with different orders of interactions. In the first normalization step,   $f({\bf{x}})$ is divided by the number of the interaction  terms ($l_i$). In the second step, we re-normalize the  fitness values to the interval $[0,1]$, this is done by subtracting the minimum fitness value among all the solutions,  and dividing by the maximum fitness value minus the minimum fitness value. 

% \subsubsection{Objectives with different order of interactions}

 Another constraint we set in some of the experiments is that, if $M_1<M_2$ then, $\beta_{U_{k_1}} = \beta_{U_{k_2}} $ for all  $|U_{k_i}|\leq M_1$. This means that all interactions contained in $f_1$ are also contained in $f_2$, but $f_2$ will also contain higher order interactions.  Starting from a single mNM-landscape $f$ of order $M$ we will generate all pairs of models $M_1,M_2$, where $M_1 \leq M_2 \leq M$. The coefficients $\beta_{U_k}$ for $f_1$ and $f_2$ will be set  as in $f$.  The idea of considering these pairs of objectives is to evaluate what is the influence in the shape of the Pareto front, and other characteristics of the MOPs, of objectives that have different order of interactions between their variables.

\section{Boltzmann distribution}

 The relationship between the fitness function and the variables dependencies that arise in the selected solutions can be modeled using the Boltzmann probability distribution  \cite{Muehlenbein_and_Mahnig:2002a,Muhlenbein_et_al:1999}.  The Boltzmann probability distribution  $p_B({\bf{x}})$ is defined as  

  \begin{equation}
  p_B({\bf{x}}) = \frac{e^{\frac{g({\bf{x}})}{T}}}{
  \sum_{{\bf{x}}'} e^{\frac{g({\bf{x}}')}{T}}}, \label{eq:BOLTPROB}
  \end{equation}
  where $g({\bf{x}})$ is a given objective function and $T$ is the system temperature that can be used as a parameter to smooth the probabilities.

The key point about $p_B({\bf{x}})$ is that it assigns a higher probability to solutions with better fitness. The solutions with the highest probability correspond to the optima.

Starting from the complete enumeration of the search space, and using as the fitness function the objectives of an  mNM-landscape, we associate to each possible solution ${\bf{x}}^i$ of the search space  $m$  probability values $(p^1_{B_i}({\bf{x}}^i), \dots, p^m_{B_i}({\bf{x}}^i))$ according to the corresponding Boltzmann probability distributions. There is one probability value for each objective and in this paper we use the same temperature parameter $T=1$ for all the distributions.

 Using the Boltzmann distribution we can investigate how potential  regularities of the fitness function are translated into statistical properties of the distribution \cite{Muehlenbein_and_Mahnig:2002a}. This question has been investigated for single-objective functions in different contexts \cite{Santana_et_al:2012h,Santana_et_al::2014a,Santana_et_al:2008a} but we have not found report on similar analysis for MOP. One relevant result in single-objective problems is that if the objective function is additively decomposable in a set of subfunctions  defined on subsets of variables (definition sets), and the definition sets satisfy certain constraints, then it is possible to factorize the associated Boltzmann distribution into a product of marginal distributions \cite{Muhlenbein_et_al:1999}. Factorizations allow problem decomposition and are at the core of EDAs.

\section{Experiments} \label{sec:EXPE}

%\subsection{Objectives}

 In our experiments we investigate the following issues:

\begin{itemize}
 \item How the parameters of mNM model determine the shape of the Pareto front?
  
 \item How is the strength of the interactions between variables influenced by the parameters of the model?

 \item Under which conditions can factorized approximations of the Boltzmann probability reproduce the shape of the Pareto front?

%  \item Evaluating the quality of alternative problem decompositions: It is possible to compute measures of divergence between $p_B({\bf{x}})$ and any possible factorization (e.g. the product of univariate marginal distributions).

\end{itemize}

%\subsection{Methodology}

Algorithm~\ref{alg:APPROACH} describes the steps of our simulations.  We use a reference NM landscape ($N=10$, $M=2$) and create a bi-objective mNM model from it using different combinations of parameters $\sigma$ and $|U_{k_i}|$.

\begin{BAlgo}{Simulation approach}
 \label{alg:APPROACH}
 \item  Define the mMN model using its parameters.  
 \item  For each objective:
 \item \T {Evaluate the $2^N$ points of the search space.}
 \item \T {Compute the Boltzmann distribution.}
 \item \T {Compute the univariate marginals from the Boltzmann distribution.}
 \item \T {For all solutions, compute univariate distribution as the product of univariate marginals.}
 \item  Determine the Pareto front using the objective values.
 \item  Determine the approximation of the Pareto front using the univariate factorizations of all the objectives.
\end{BAlgo}

\subsection{Influence of the mNM-landscape parameters}

We investigate how the parameters of mNM model determine the shape of the Pareto.

%We begin this section by presenting one illustrative example of the kind of analysis that can be done with the combination of the nNM landscape and probabilistic modeling using the Boltzmann distribution. 

\begin{figure*}[htbp]
 \begin{center}

\includegraphics[width=2.00in]{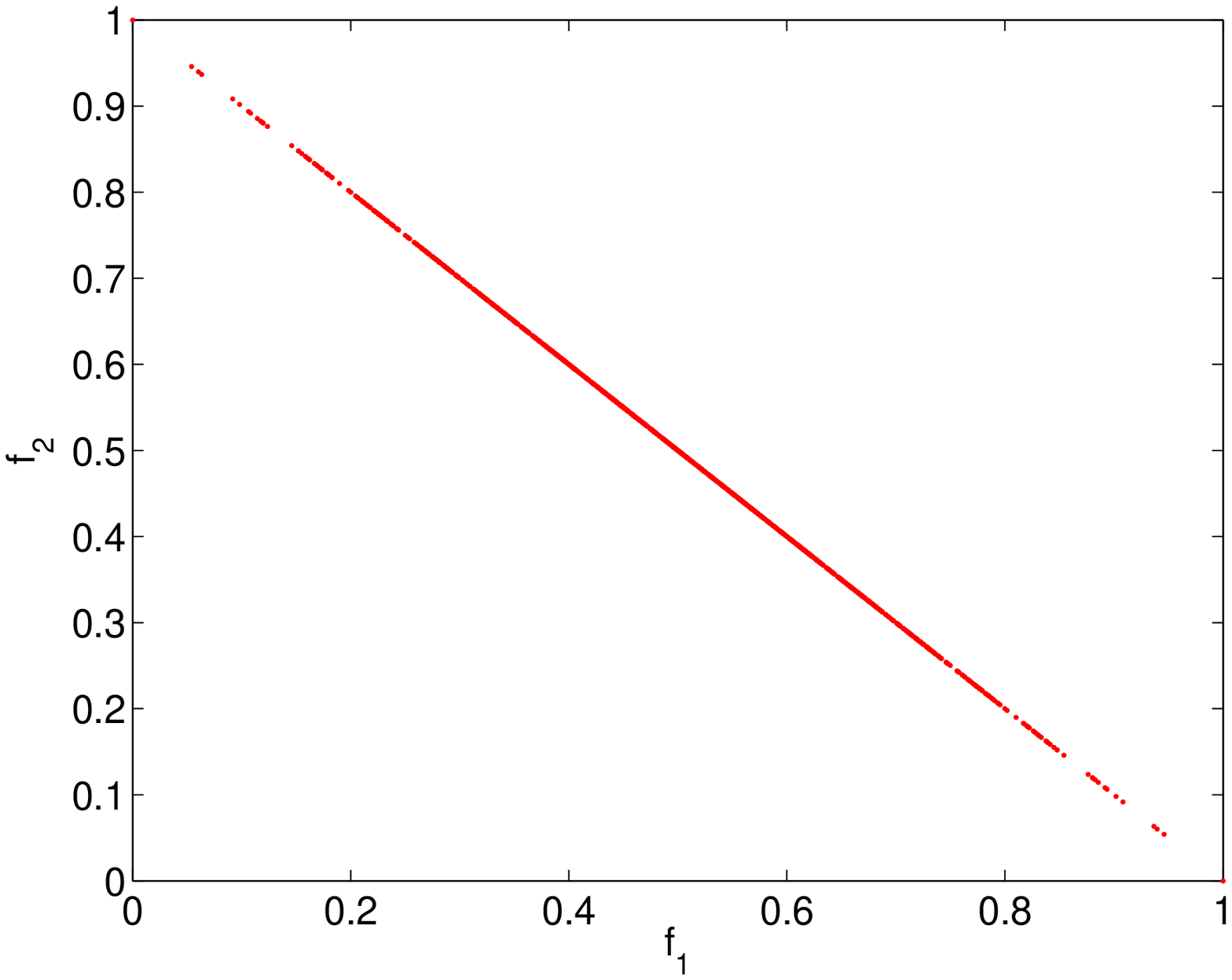}
\includegraphics[width=2.00in]{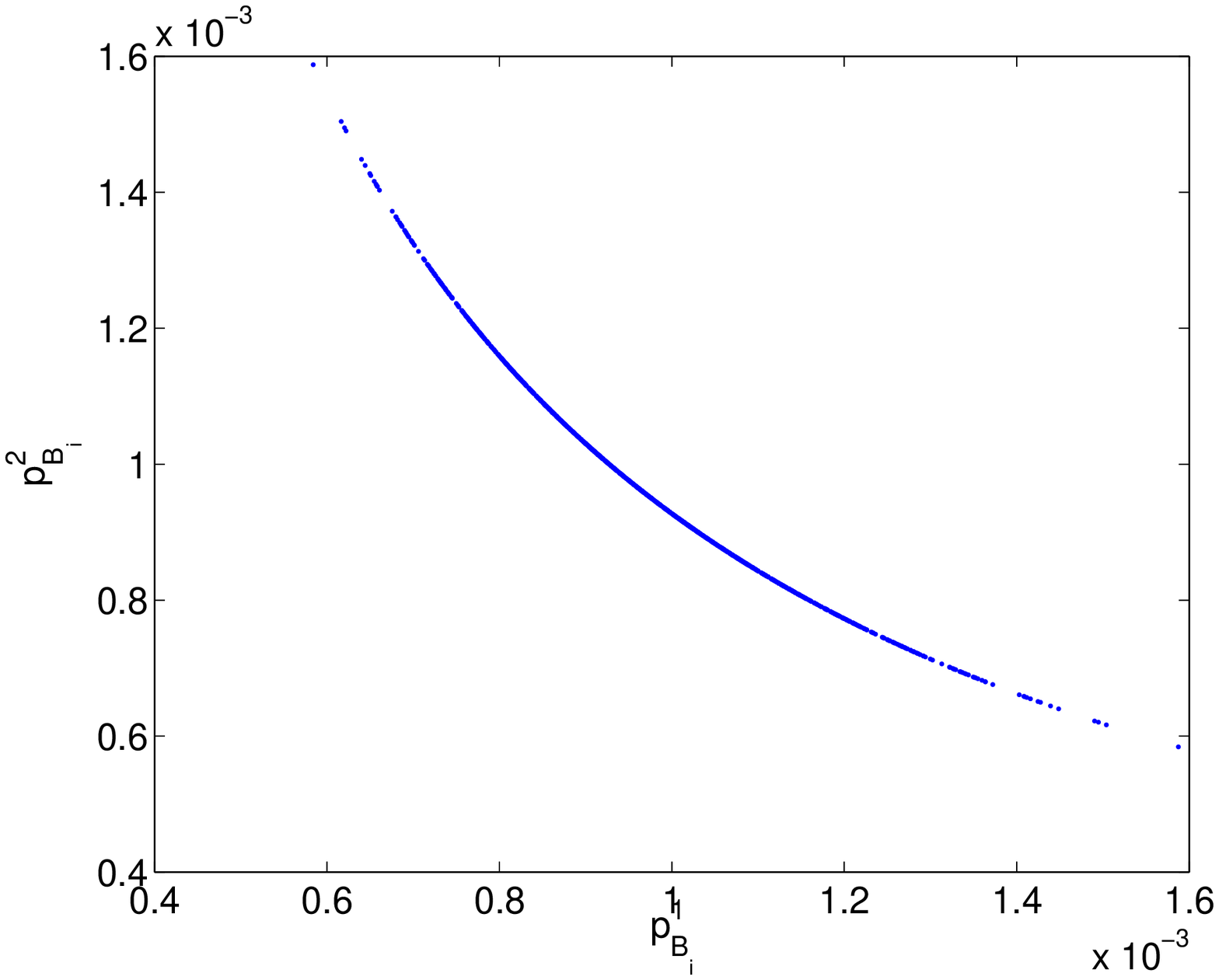}
  \includegraphics[width=2.00in]{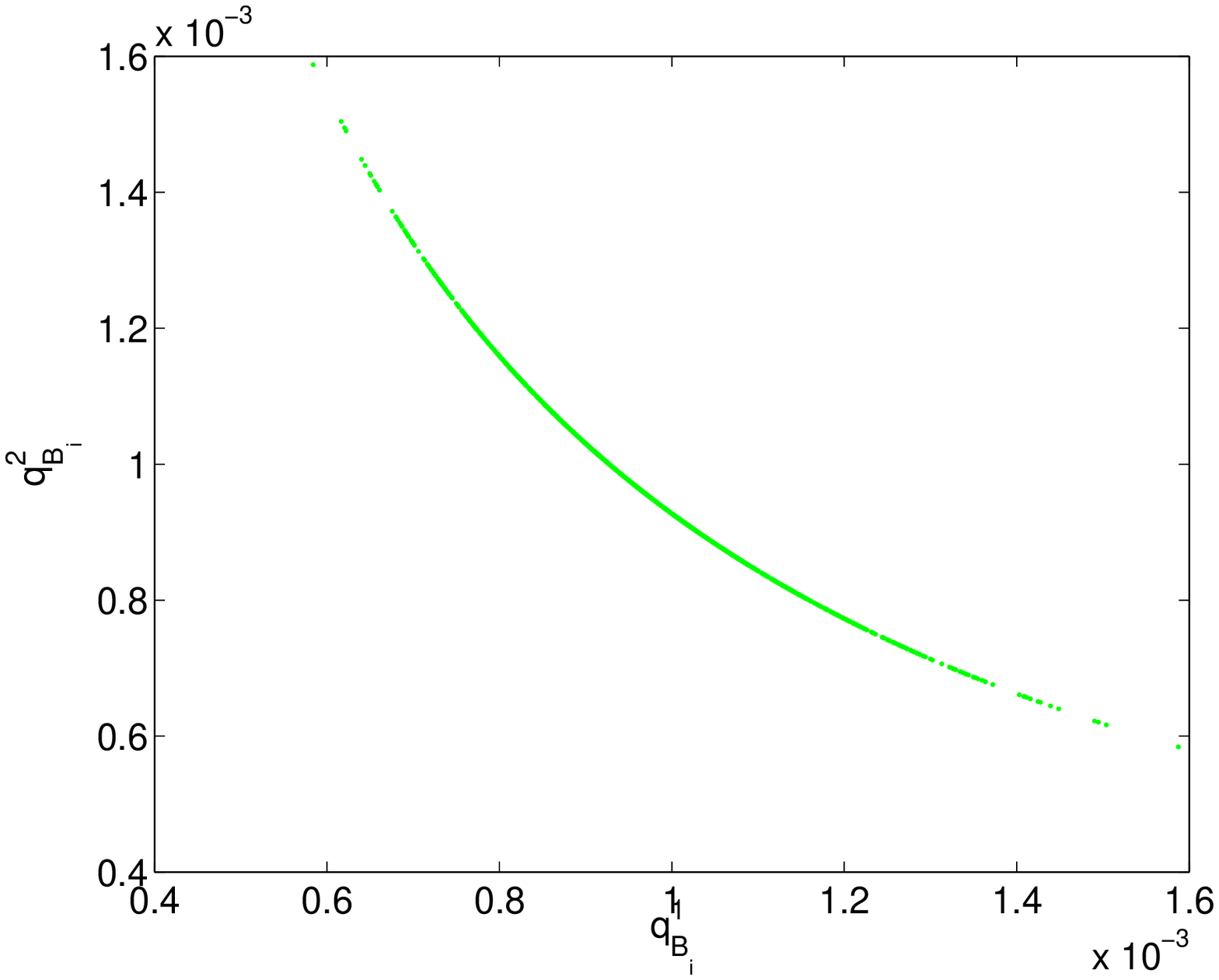}

\includegraphics[width=2.00in]{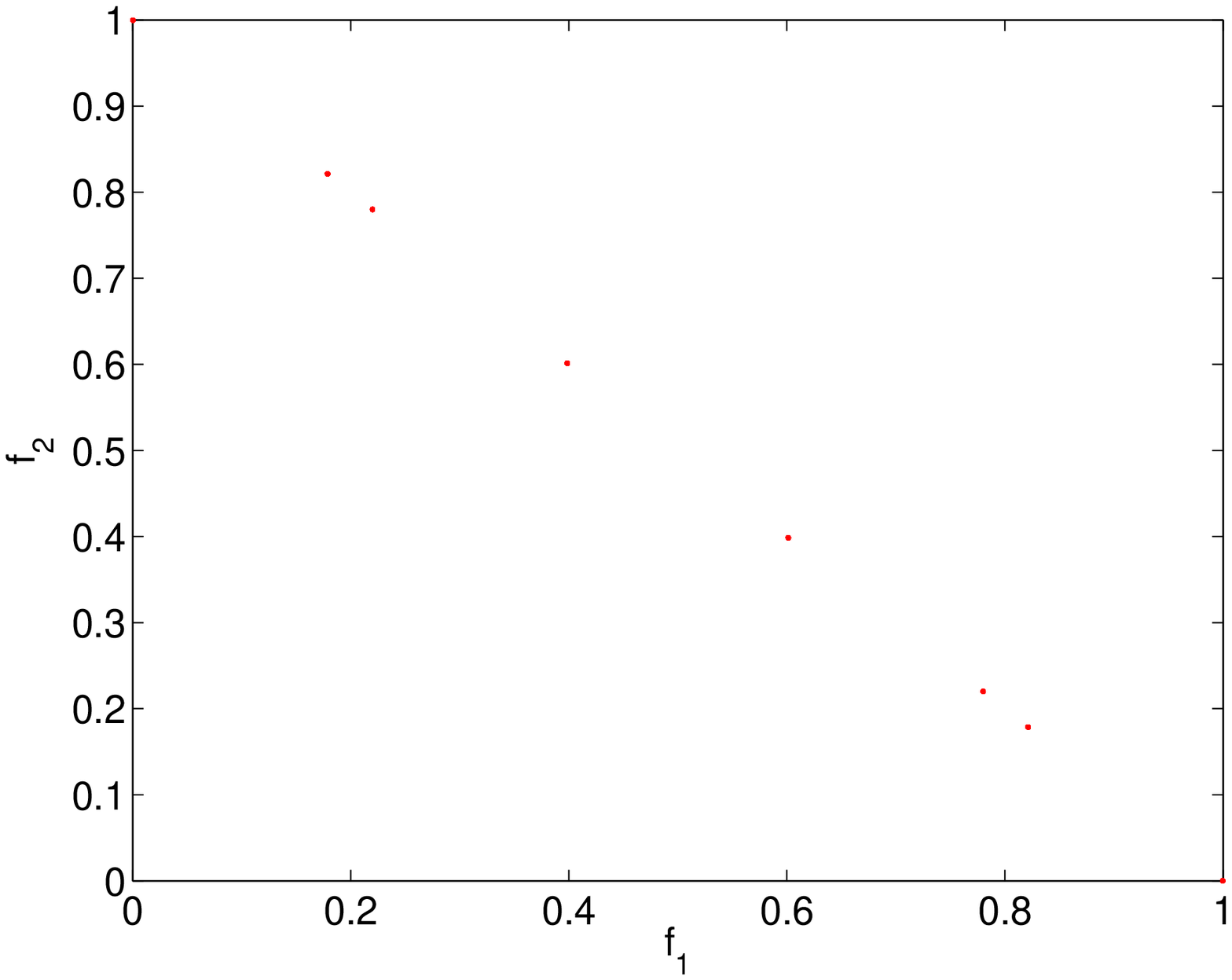}
\includegraphics[width=2.00in]{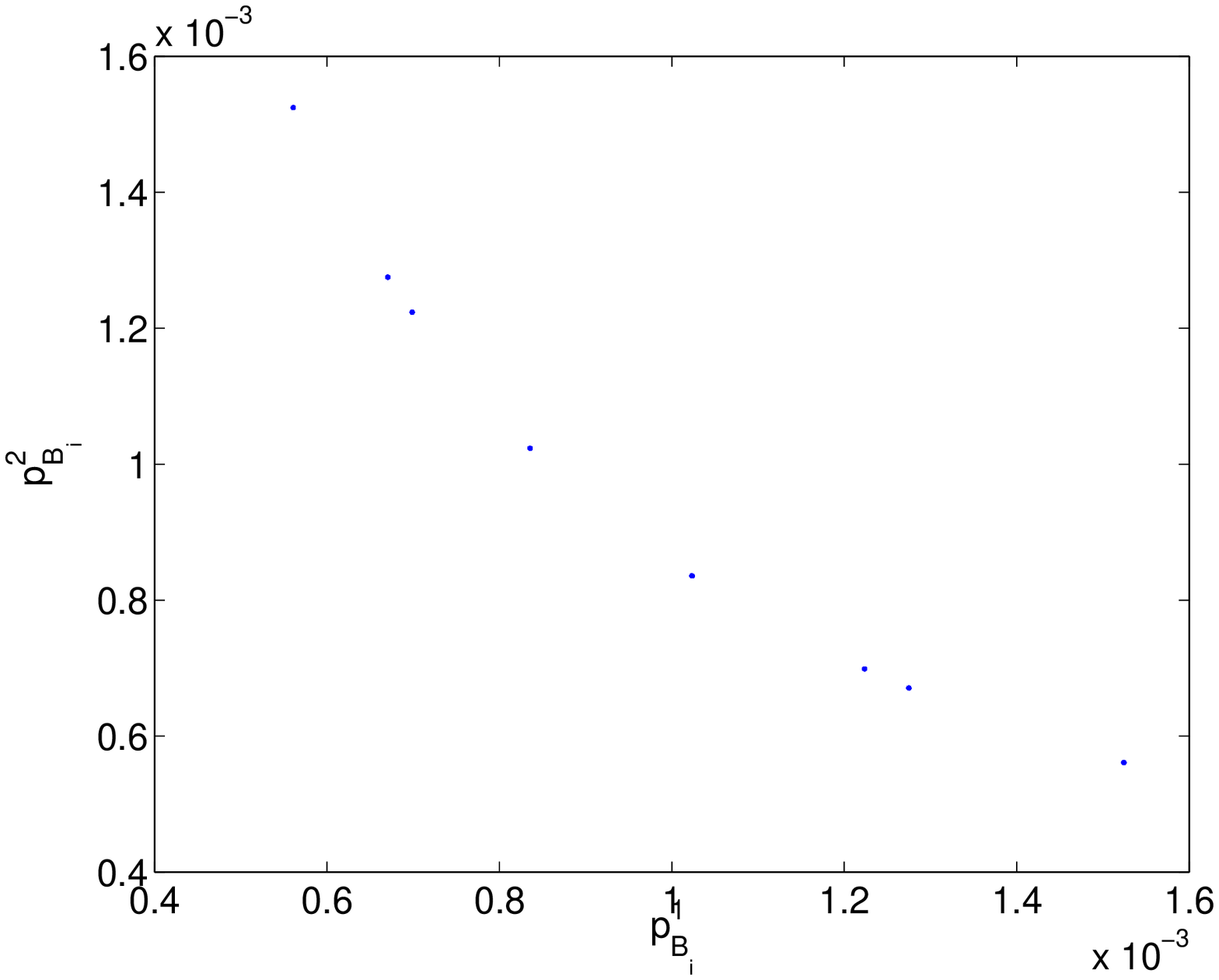}
  \includegraphics[width=2.00in]{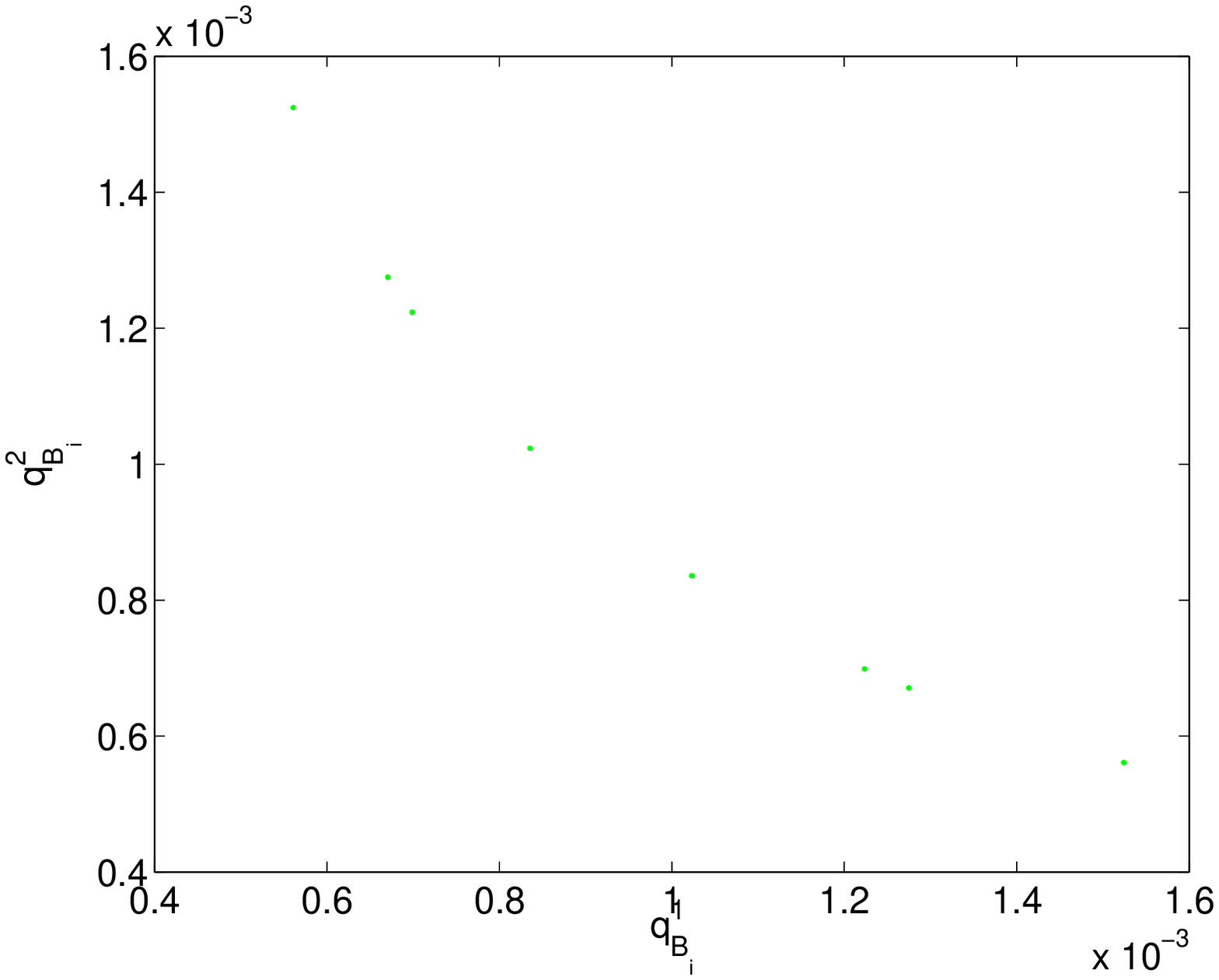}

\includegraphics[width=2.00in]{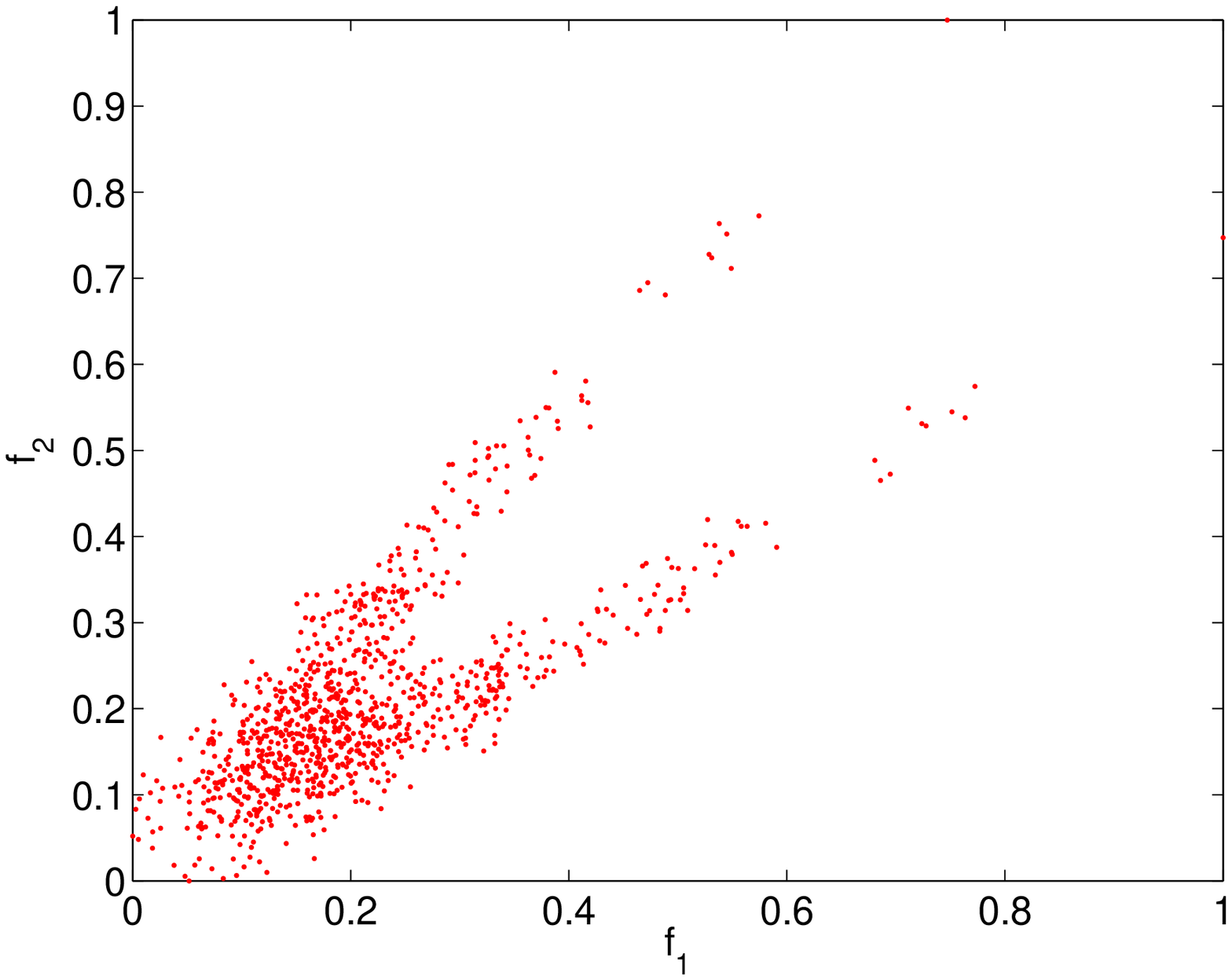}
\includegraphics[width=2.00in]{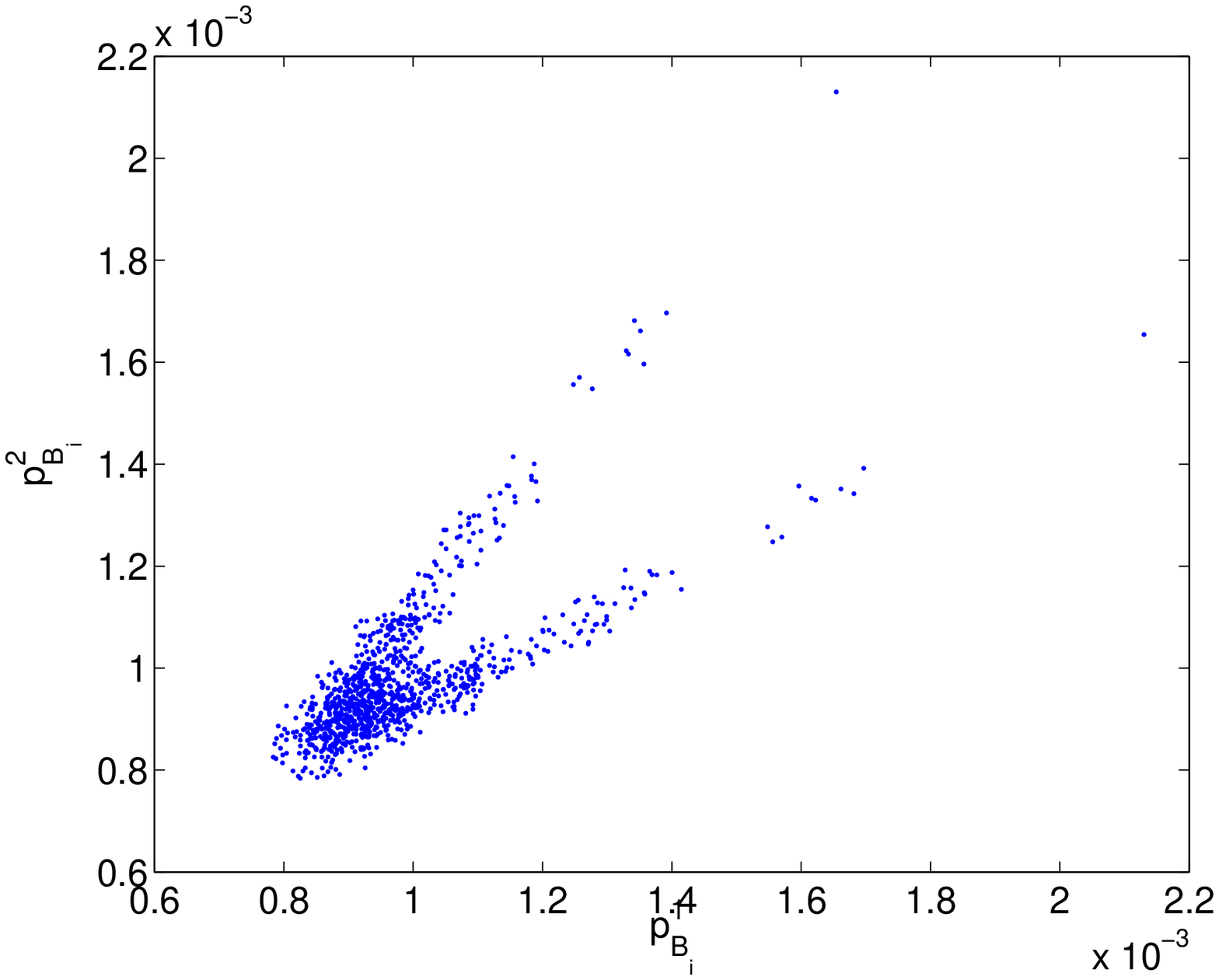}
  \includegraphics[width=2.00in]{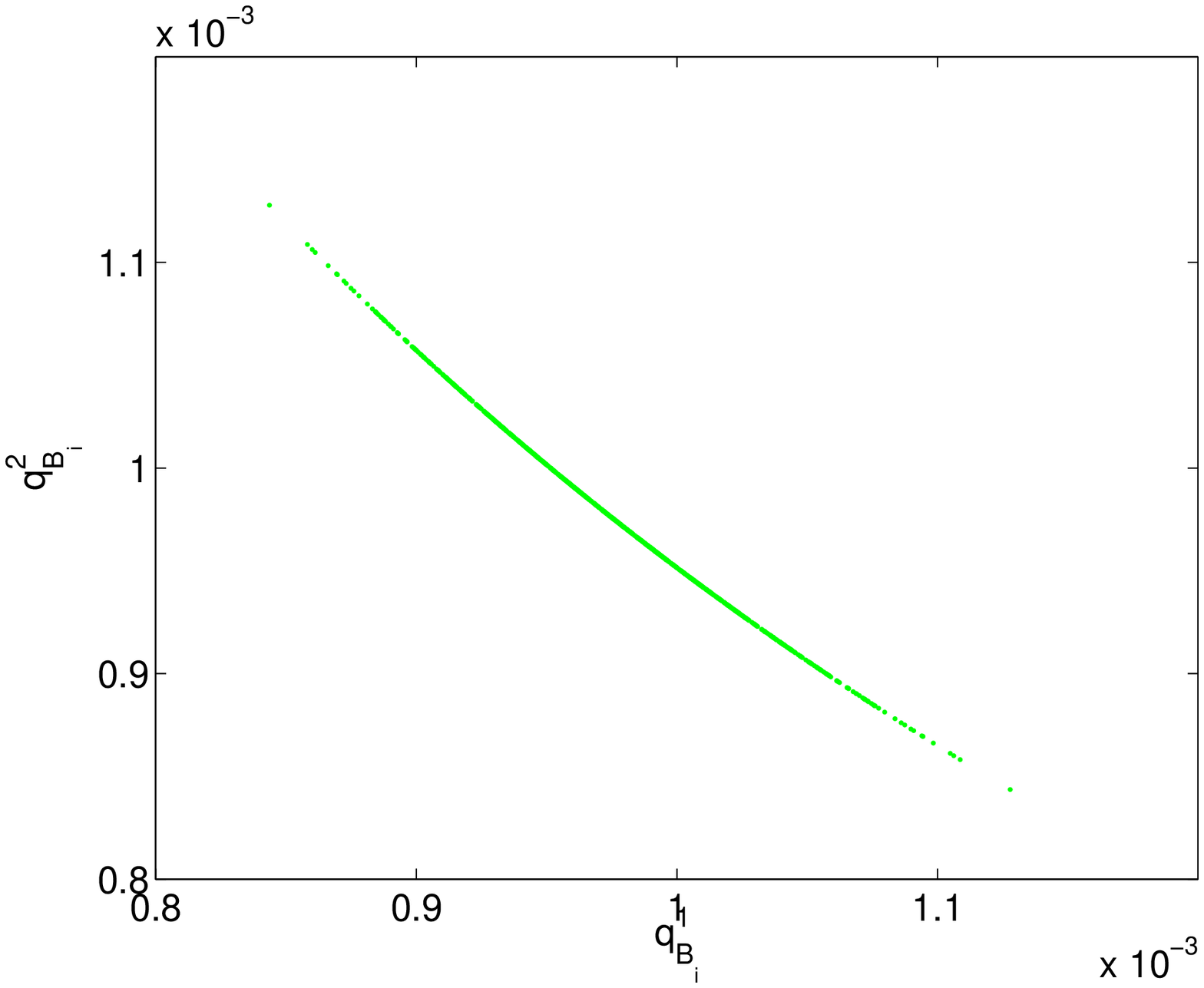}

\includegraphics[width=2.00in]{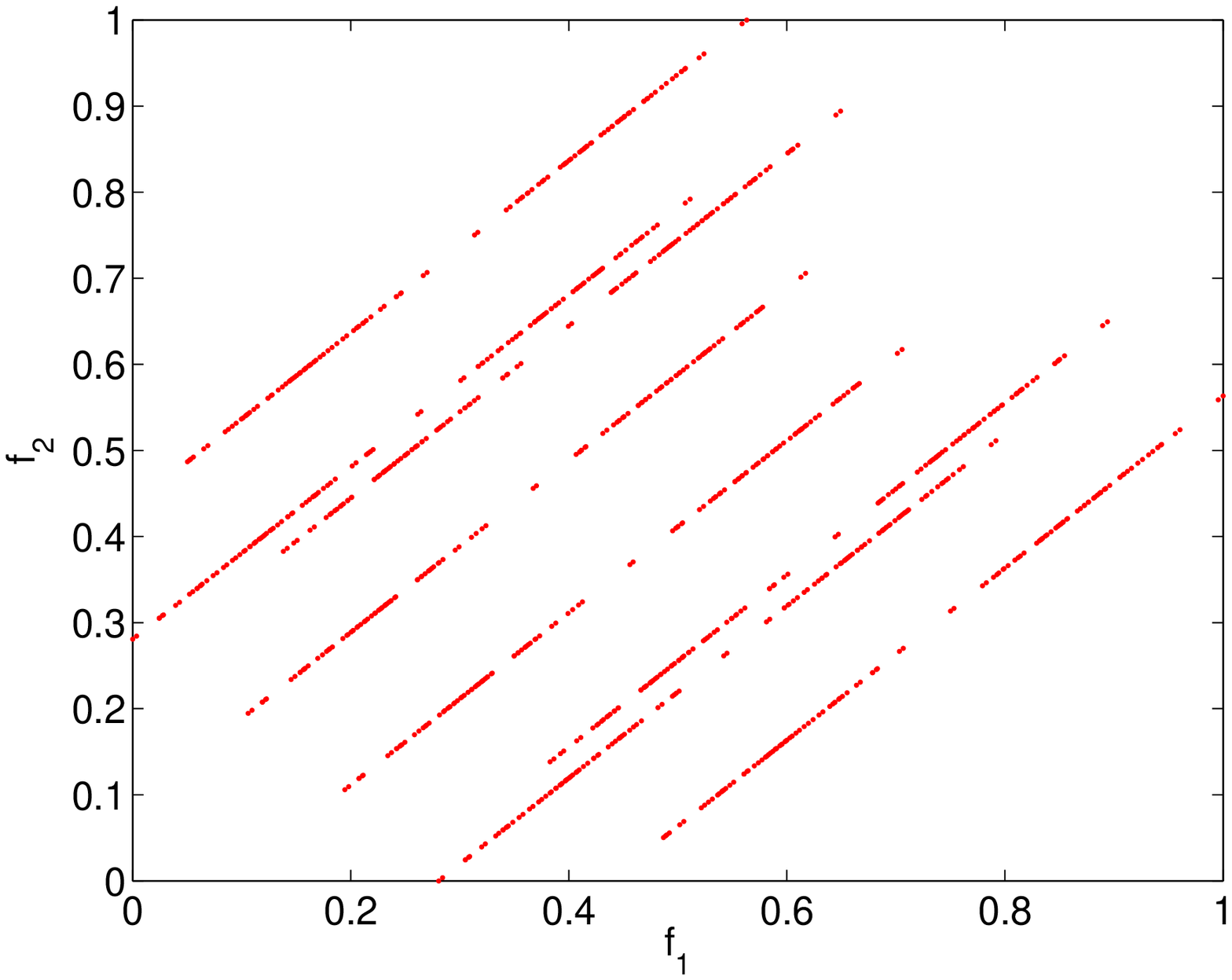}
\includegraphics[width=2.00in]{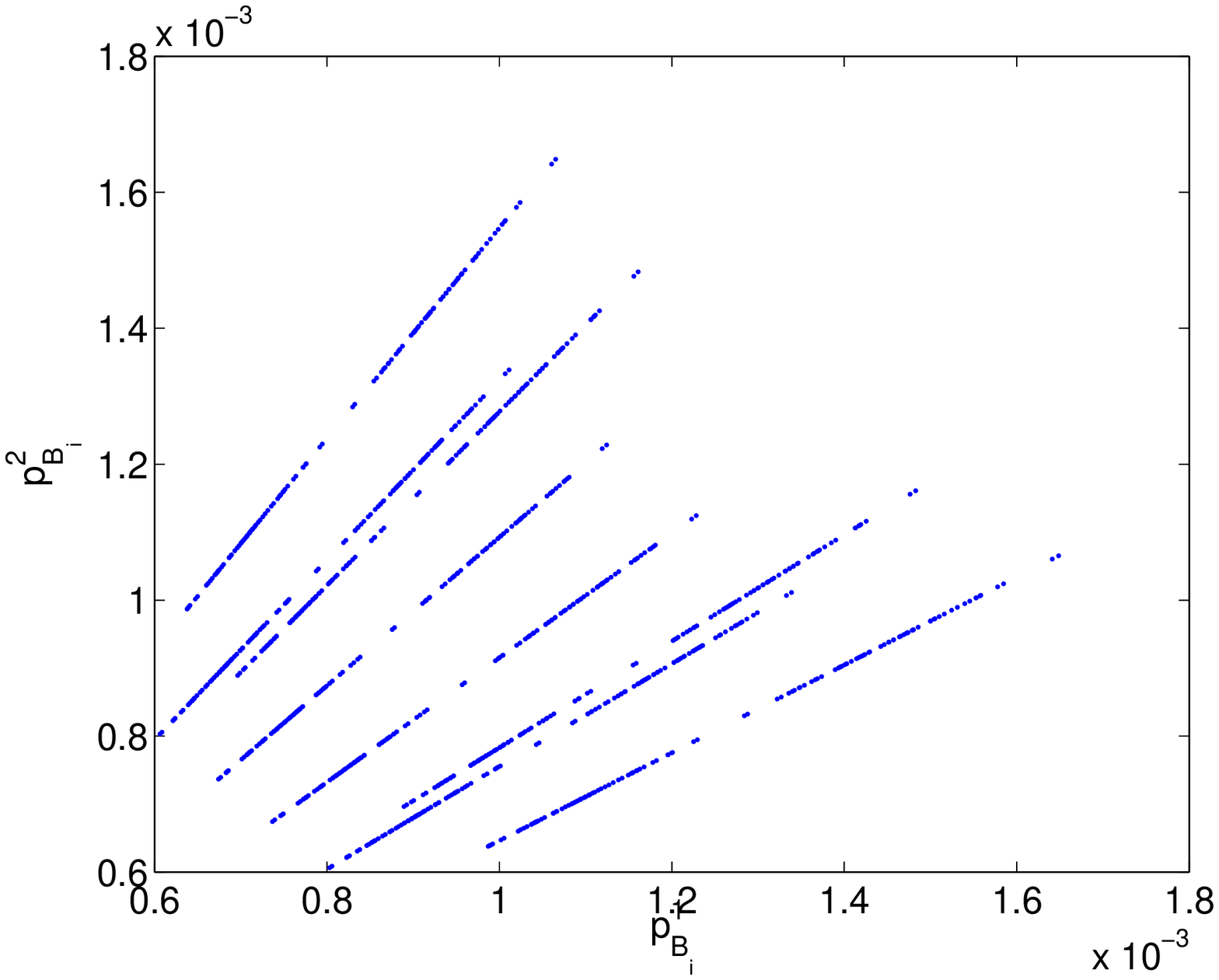}
  \includegraphics[width=2.00in]{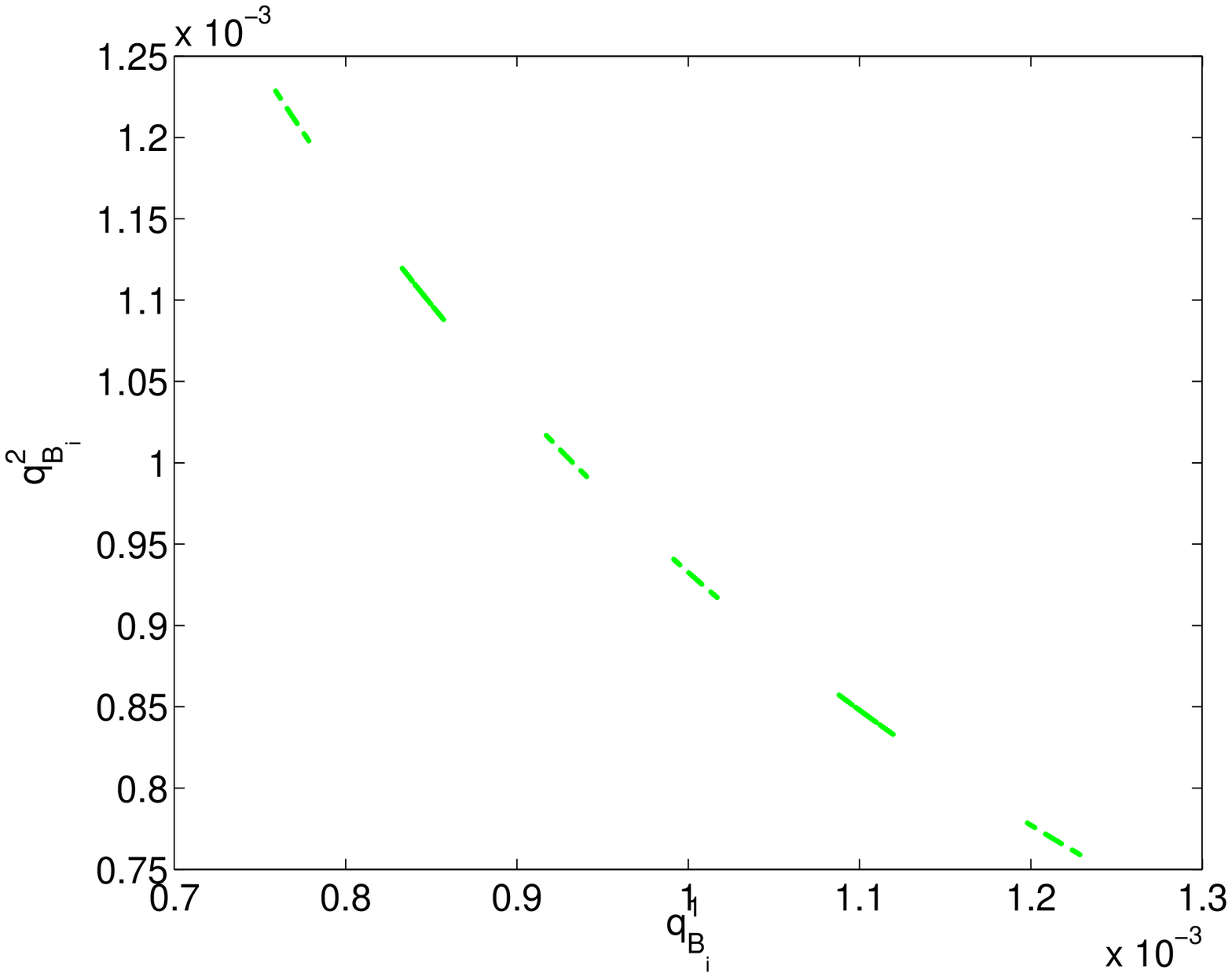}
  \caption{Objective values, Boltzmann distribution, and univariate approximations for different values of $\sigma$ and different maximum orders of interactions. Column 1: Evaluation of the $2^{10}$  solutions  that are part of the search space for $N=10$ and different values of $\sigma$ and $|U_{k_i}|$.  From row 1 to row 4, the figures respectively show the objective values of the mNM model for ($\sigma=1, |U_{k_i}|=1$), ($\sigma=19, |U_{k_i}|=1$), ($\sigma=1, |U_{k_i}|=2$) and ($\sigma=19, |U_{k_i}|=2$). Column 2: Boltzmann distributions computed from the objectives. Column 3: Univariate approximations of the Boltzmann distributions. } 
  \label{fig:mNM_MOD}
 \end{center}
\end{figure*}

Figure~\ref{fig:mNM_MOD} (column 1)  shows the evaluation of the $2^{10}$  solutions  that are part of the search space for $N=10$ and different values of $\sigma$ and $|U_{k_i}|$.  From row 1 to row 4, the figures respectively show the objective values of the mNM landscape for different combination of its parameters: ($\sigma=1, |U_{k_i}|=1$), ($\sigma=1, |U_{k_i}|=19$), ($\sigma=2, |U_{k_i}|=1$), ($\sigma=2, |U_{k_i}|=19$).   

The influence of $\sigma$ can be seen by comparing the figure in row 1 with the figure in row 2, and doing a similar comparison with figures in row 3 and row 4.  Increasing $\sigma$ from $1$ to $19$ produces a clustering of the points in the objective space. One reason for this behavior is that several genotypes will map to the same objective values. The clustering effect in the space of objectives is a direct result of the clumpiness effect described for the NM-model when $\sigma$ is increased \cite{Manukyan_et_al:2014}.

The effect of the maximum order of the interactions can be seen by comparing the figure in row 1 with the figure in row 3, and the figures in rows 2 and 4. For $\sigma=1$, adding interactions transforms the shape of the Pareto front from a line to a boomerang-like shape. For $\sigma=19$, the $8$ points are transformed into a set of $8$ stripes that seem to be parallel to each other. In both cases, the changes due to the increase in the order of the interactions are remarkable.

\begin{figure*}[htbp]
 \begin{center}
  \includegraphics[width=3.0in]{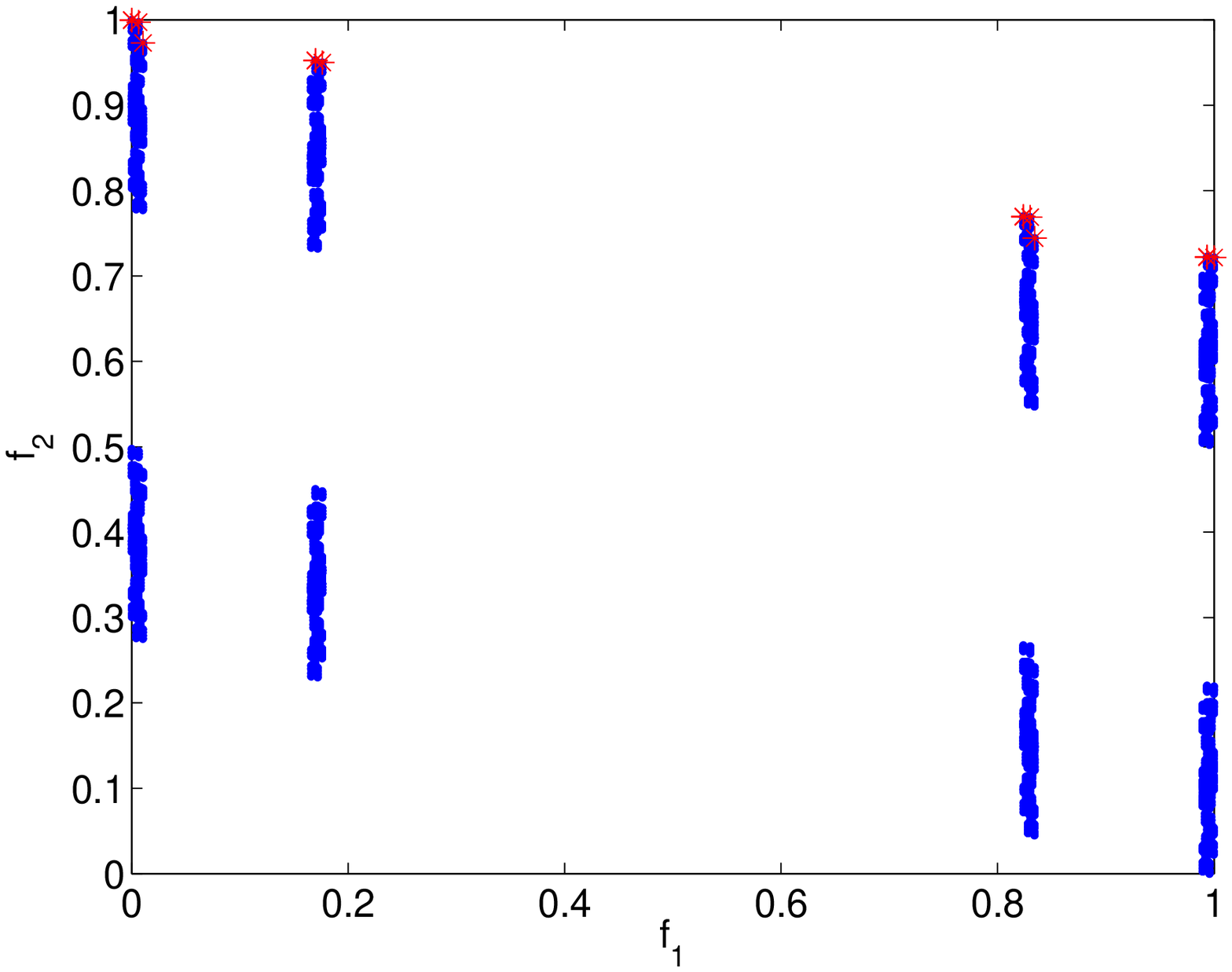}
  \includegraphics[width=3.0in]{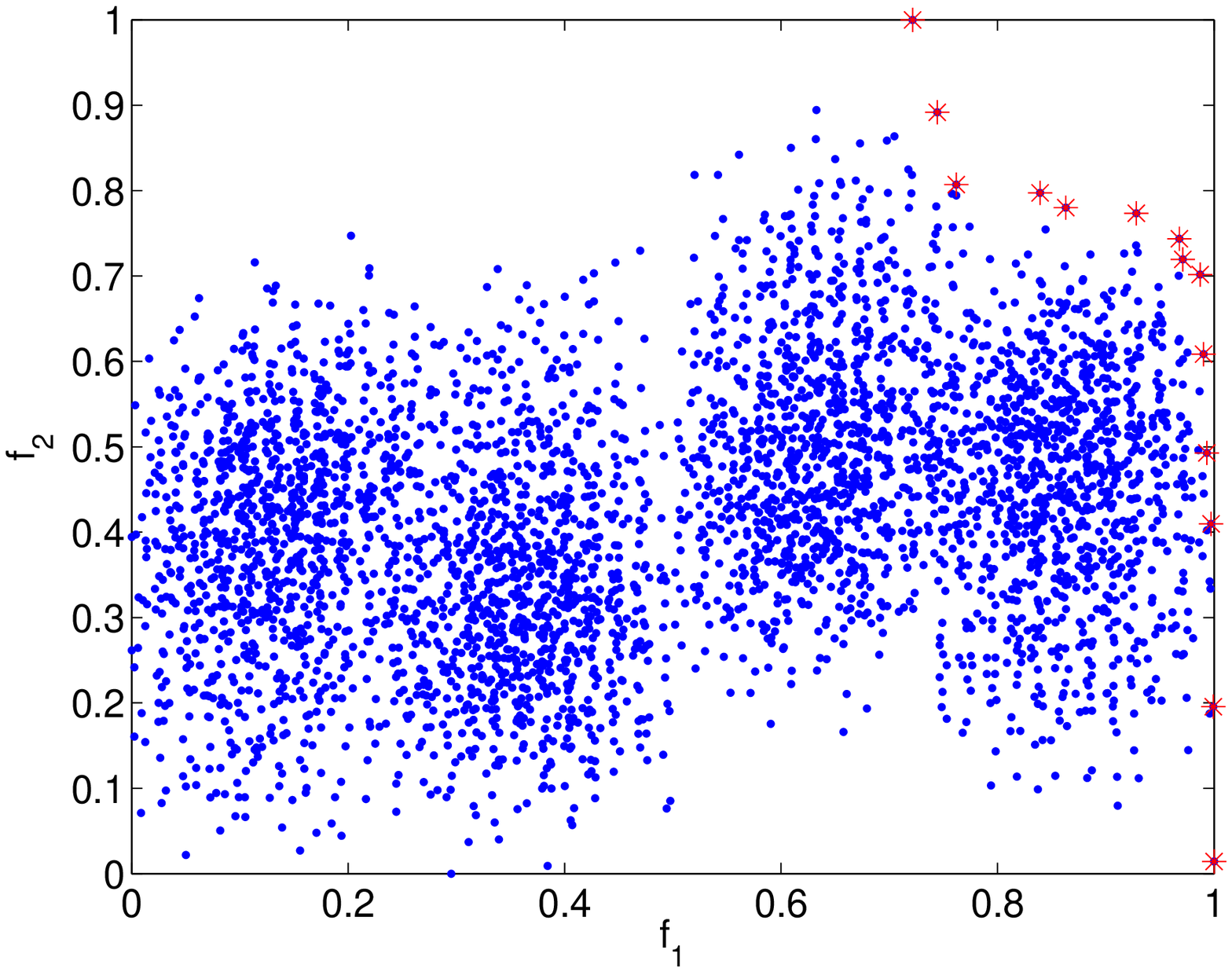} 
  \caption{Objective values and Pareto fronts of the mNM model for $\sigma=36$ and different maximum orders of interactions: Left) $|U_{k_1}|=1$ and $|U_{k_2}|=2$, Right) $|U_{k_1}|=2$ and $|U_{k_2}|=3$.}
  \label{fig:ConPF_sigma36}
 \end{center}
\end{figure*}

In the next experiments, and in order to emphasize the flexibility of the mNM-landscape, we allow the two objectives of the same mNM-landscape to have different maximum order of interactions. Figure~\ref{fig:ConPF_sigma36} shows the objective values and Pareto fronts of the mNM model for $\sigma=36$ for the situation in which $f_1$ has a maximum order of interactions $|U_{k_1}|=o$ and  $f_2$ has a maximum order of interactions $|U_{k_2}|=o+1$. It can be observed that the shapes of the fronts are less regular than in the previous experiments but some regularities are kept.

\subsection{Boltzmann distribution}

 Figure~\ref{fig:mNM_MOD} (column 2) shows the Boltzmann probabilities associated to each mNM-landscape model described in column 1, i.e., $(p^1_{B_i}({\bf{x}}^i),p^2_{B_i}({\bf{x}}^i))$. 

   The Boltzmann distribution modifies the shape of the objective space but it does not modify the solutions that belong to the Pareto set. This is so  because the dominance relationships between the points are preserved by the Boltzmann distribution. However, the Boltzmann distribution ``bends'' the original objective space. This effect can be clearly appreciated in rows 1 and row 4. In the first case, the line is transformed into a curve. In the second case the parallel lines stripes that appear in the original objective space change direction. 

 The Boltzmann distribution can be used as an effective way to modify the shape of the Pareto while keeping the dominance relationships. This can be convenient to modify the spacing between Pareto-optimal solutions,  for more informative visualization of the objective space, and for investigating how changes in the strength of selection could be manifested in the shape of the Pareto front approximations.

\subsection{Factorized univariate approximations}

Figure~\ref{fig:mNM_MOD} (column 3) shows the approximations of the Boltzmann distributions for the two objectives, each approximation computed using the corresponding product of the univariate marginals, i.e., $(q^1_{B_i}({\bf{x}}^i),q^2_{B_i}({\bf{x}}^i))$. For $|U_{k}|=1$, the approximations are identical to the Boltzmann distribution. This is because the Boltzmann distribution can be exactly factorized in the product of its univariate marginal distributions. Therefore, as a straightforward extension of the factorization theorems available for the single-objective additive functions, we hypothesize that if the structure of all objectives is decomposable and the decompositions satisfy the running intersection property  \cite{Muehlenbein_and_Mahnig:2002a,Muhlenbein_et_al:1999}, then \emph{the associated factorized distributions will preserve the shape of the Pareto front}.   

 However, the univariate approximation does not always respect the dominance relationships and this fact provokes changes in the composition and shape of the Pareto front. This can be appreciated in rows 3 and 4, where the univariate approximation clearly departs from the Boltzmann distribution. Still, as shown in row 4, some characteristics of the original function, as the discontinuity in the space of objectives, can hold for the univariate factorization. 

An open question is under which conditions will the univariate approximation keep the dominance relationship between the solutions. One conjecture is that if the factorized approximation keeps the ranking of the original functions for all the objectives then the dominance relationship will be kept, but this condition may not be necessary. The answer to this question is beyond the scope of this paper. Nevertheless, we include the discussion to emphasize why explicit modeling of interactions by means of the mNM landscape together with the use of the Boltzmann distribution is relevant for the study of MOPs.

\subsection{Interactions and dependencies in the mNM landscape}

 By computing bivariate and univariate marginals from the Boltzmann distribution and computing the mutual information for every pair of variables we can assess which are the strongest pair-wise interactions.

\begin{figure}[htbp]
 \begin{center}
  \includegraphics[width=3.5in]{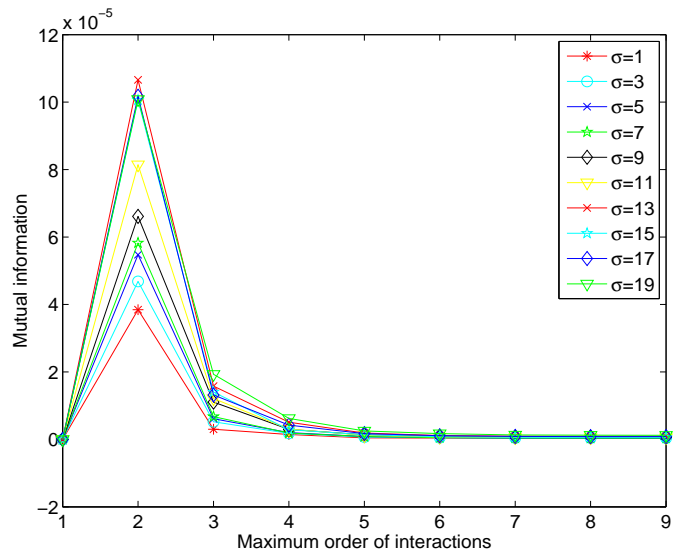}
   \caption{Influence of the maximum order of the interactions and $\sigma$ in the mutual information.}
  \label{fig:MUTINF}
 \end{center}
\end{figure}

  In this section we analyze how the maximum order of the interactions and the $\sigma$ parameter affect the dependencies in the Boltzmann distribution. A reference NM model with ($N=10$) was generated and by varying the  parameters $M \in \{1,\dots,9\}$ and  $\sigma=2i+1, i \in \{0,1,\dots,9\}$) we generated different mNM landscapes. The results presented in this section are the average of $10$ models for each combination of parameters.  We focus on the analysis of the dependencies in only one of the objectives.  

Figure~\ref{fig:MUTINF} shows the values of the mutual information for the combinations of the maximum order of the interactions and $\sigma$. When the maximum order of the interactions is $1$, the approximation given by the univariate factorization is exact, therefore, the mutual information between the variables are $0$ for all values of $\sigma$. The  mutual information is maximized when the maximum number of interactions is $2$. For these mNM landscapes we would expect the univariate approximation to considerably distort the shape of the Pareto front, as shown in Figure~\ref{fig:mNM_MOD}, column 3, rows 3 and 4. 

%Interestingly, the influence that adding interactions has in the mutual information values is similar to the effect of increasing the number of unitation constraints on the mutual information of independent variables \cite{Santana_et_al:2001c}.

Figure~\ref{fig:MUTINF} shows that $\sigma$ can be used to tune the strength of the interactions between the variables. As $\sigma$ increases the mutual information also increases. This fact would allow us to define objectives that have interactions of the same maximum order but with different strength.

\subsection{Discussion}
 
  We summarize some of the findings from the experiments:

\begin{itemize} 
  \item Univariate factorizations are poor approximations for mNM models of maximum order two and higher. 
  \item The mutual information between the variables of the NM landscape is maximized for problems with maximum order of interaction $2$.
  \item The $\sigma$ parameter can be used for changing the shape of the Pareto fronts and increasing the strength of the interactions in the objectives. In particular, there is a direct effect of $\sigma$ in the discontinuity of the Pareto front and the emergence of clusters.   
\end{itemize}

\section{Conclusions} \label{sec:CONCLU}
   
We have shown how the mNM landscape  can be used to investigate the effect that interactions between the variables have in the shapes of the fronts and in the emergence of dependencies between the variables of the problem. We have shown that the Boltzmann distribution can be used in conjunction with the mNM model to investigate how interactions are translated into dependencies. A limitation of the Boltzmann distribution is that is can be computed exactly only for problems of limited size. 

 The idea of using the Boltzmann distribution to modify the Pareto shape of the functions can be related to previous work by Okabe et al. \cite{Okabe_et_al:2004a}  on the application of deformations, rotations, and shift operators to generate test functions with difficult Pareto sets. However, by using the Boltzmann distribution we explicitly relate the changes in  the shape of the Pareto to the relationship interactions-dependencies determined by the Boltzmann distribution. This can be considered as an alternative path to other approaches to creation of benchmarks for MOPs, like the combination of single-objectives functions of known difficulty \cite{Brockhoff_et_al:2015} or the maximization of problem difficulty by applying direct optimization approaches \cite{Santana_et_al:2015}. 

 Our results can be useful for the conception and validation of MOEAs that use probabilistic modeling. In this direction, we have advanced the idea that the effectiveness of a factorized approximation in the context of MOPs may be related to the way it preserves the original dominance relationships between solutions. We have shown that the Boltzmann distribution changes the shape of the fronts but does not change which solutions belong to the Pareto front.

\section*{Acknowledgment}

This work has been partially supported by IT-609-13 program (Basque Government) and the TIN2013-41272P (Spanish Ministry of Science and Innovation) project.  R. Santana acknowledges support from the Program Science Without Borders  No. : 400125/2014-5).

\bibliographystyle{abbrv}

%\bibliography{../../NewThesis/Thesbib}

\begin{thebibliography}{10}

\bibitem{Aguirre_and_Tanaka:2004}
H.~Aguirre and K.~Tanaka.
\newblock Insights and properties of multiobjective {MNK}-landscapes.
\newblock In {\em Proceedings of the 2004 Congress on Evolutionary Computation
  CEC-2004}, pages 196--203, Portland, Oregon, 2004. IEEE Press.

\bibitem{Aguirre_and_Tanaka:2007}
H.~E. Aguirre and K.~Tanaka.
\newblock Working principles, behavior, and performance of {MOEAs} on
  {MNK}-landscapes.
\newblock {\em European Journal of Operational Research}, 181(3):1670--1690,
  2007.

\bibitem{Bosman_and_Thierens:2002}
P.~A. Bosman and D.~Thierens.
\newblock Multi-objective optimization with diversity preserving mixture-based
  iterated density estimation evolutionary algorithms.
\newblock {\em International Journal of Approximate Reasoning}, 31(3):259--289,
  2002.

\bibitem{Brockhoff_et_al:2015}
D.~Brockhoff, T.-D. Tran, and N.~Hansen.
\newblock Benchmarking numerical multiobjective optimizers revisited.
\newblock In {\em Proceedings of the Companion Publication of the 2015 on
  Genetic and Evolutionary Computation Conference}, pages 639--646, Madrid,
  Spain, 2015.

\bibitem{Kauffman:1993}
S.~Kauffman.
\newblock {\em Origins of Order}.
\newblock Oxford University Press, 1993.

\bibitem{Larranaga_et_al:2012}
P.~Larra{\~{n}}aga, H.~Karshenas, C.~Bielza, and R.~Santana.
\newblock A review on probabilistic graphical models in evolutionary
  computation.
\newblock {\em Journal of Heuristics}, 18(5):795--819, 2012.

\bibitem{Lopez_et_al:2014}
M.~L{\'o}pez-Ib{\'a}nez, A.~Liefooghe, and S.~Verel.
\newblock Local optimal sets and bounded archiving on multi-objective
  {NK}-landscapes with correlated objectives.
\newblock In {\em Parallel Problem Solving from Nature--PPSN XIII}, pages
  621--630. Springer, 2014.

\bibitem{Lozano_et_al:2005}
J.~A. Lozano, P.~Larra{\~{n}}aga, I.~Inza, and E.~Bengoetxea, editors.
\newblock {\em Towards a {N}ew {E}volutionary {C}omputation: {A}dvances on
  {E}stimation of {D}istribution {A}lgorithms}.
\newblock Springer, 2006.

\bibitem{Manukyan_et_al:2014}
N.~Manukyan, M.~J. Eppstein, and J.~S. Buzas.
\newblock Tunably rugged landscapes with known maximum and minimum.
\newblock {\em arXiv.org}, arXiv:1409.1143, 2014.

\bibitem{Marti_et_al:2010}
L.~Marti, J.~Garcia, A.~Berlanga, C.~A. Coello, and J.~M. Molina.
\newblock On current model-building methods for multi-objective estimation of
  distribution algorithms: Shortcommings and directions for improvement.
\newblock Technical Report GIAA2010E001, Department of Informatics of the
  Universidad Carlos III de Madrid, Madrid, Spain, 2010.

\bibitem{Muehlenbein_and_Mahnig:2002a}
H.~M{\"u}hlenbein and T.~Mahnig.
\newblock Evolutionary algorithms and the {B}oltzmann distribution.
\newblock In K.~A. DeJong, R.~Poli, and J.~Rowe, editors, {\em Foundation of
  Genetic Algorithms 7}, pages 133--150. Morgan Kaufmann, 2002.

\bibitem{Muhlenbein_et_al:1999}
H.~M{\"{u}}hlenbein, T.~Mahnig, and A.~Ochoa.
\newblock Schemata, distributions and graphical models in evolutionary
  optimization.
\newblock {\em Journal of Heuristics}, 5(2):213--247, 1999.

\bibitem{Okabe_et_al:2004a}
T.~Okabe, Y.~Jin, M.~Olhofer, and B.~Sendhoff.
\newblock On test functions for evolutionary multi-objective optimization.
\newblock In {\em Parallel Problem Solving from Nature-PPSN VIII}, pages
  792--802. Springer, 2004.

\bibitem{Pelikan_et_al:2006a}
M.~Pelikan, K.~Sastry, and D.~E. Goldberg.
\newblock Multiobjective estimation of distribution algorithms.
\newblock In M.~Pelikan, K.~Sastry, and E.~Cant\'u-Paz, editors, {\em Scalable
  Optimization via Probabilistic Modeling: From Algorithms to Applications},
  Studies in Computational Intelligence, pages 223--248. Springer, 2006.

\bibitem{Santana_et_al:2012h}
R.~Santana, C.~Bielza, and P.~Larra{\~n}aga.
\newblock Conductance interaction identification by means of {B}oltzmann
  distribution and mutual information analysis in conductance-based neuron
  models.
\newblock {\em BMC Neuroscience}, 13(Suppl 1):P100, 2012.

\bibitem{Santana_et_al:2005b}
R.~Santana, P.~Larra{\~{n}}aga, and J.~A. Lozano.
\newblock Interactions and dependencies in estimation of distribution
  algorithms.
\newblock In {\em Proceedings of the 2005 Congress on Evolutionary Computation
  CEC-2005}, pages 1418--1425, Edinburgh, U.K., 2005. IEEE Press.

\bibitem{Santana_et_al:2008a}
R.~Santana, P.~Larra{\~{n}}aga, and J.~A. Lozano.
\newblock Protein folding in simplified models with estimation of distribution
  algorithms.
\newblock {\em IEEE Transactions on Evolutionary Computation}, 12(4):418--438,
  2008.

\bibitem{Santana_et_al::2014a}
R.~Santana, R.~B. McDonald, and H.~G. Katzgraber.
\newblock A probabilistic evolutionary optimization approach to compute
  quasiparticle braids.
\newblock In {\em Proceedings of the 10th International Conference Simulated
  Evolution and Learning (SEAL-2014)}, pages 13--24. Springer, 2014.

\bibitem{Santana_et_al:2015}
R.~Santana, A.~Mendiburu, and J.~A. Lozano.
\newblock Evolving {MNK}-landscapes with structural constraints.
\newblock In {\em Proceedings of the {IEEE} Congress on Evolutionary
  Computation {CEC} 2015}, pages 1364--1371, Sendai, Japan, 2015. IEEE press.

\bibitem{Santana_et_al:2015b}
R.~Santana, A.~Mendiburu, and J.~A. Lozano.
\newblock Multi-objective {NM}-landscapes.
\newblock In {\em Proceedings of the Companion Publication of the 2015 on
  Genetic and Evolutionary Computation Conference}, pages 1477--1478, Madrid,
  Spain, 2015.

\bibitem{Verel_et_al:2013}
S.~Verel, A.~Liefooghe, L.~Jourdan, and C.~Dhaenens.
\newblock On the structure of multiobjective combinatorial search
  space:{MNK}-landscapes with correlated objectives.
\newblock {\em European Journal of Operational Research}, 227(2):331--342,
  2013.

\end{thebibliography}

% that's all folks
\end{document}